\documentclass[letterpaper, 10 pt, conference]{ieeeconf}  

\IEEEoverridecommandlockouts                              

\usepackage{graphicx} 
\usepackage{amsmath} 
\usepackage{amssymb}  
\usepackage{cite}
\usepackage{caption}
\usepackage{subcaption}
\usepackage{hyperref}

\title{\LARGE \bf
Directional Compliance in Obstacle-Aided Navigation for Snake Robots}

\author{Tianyu Wang$^{1}$, Julian Whitman$^{1}$, Matthew Travers$^{2}$, Howie Choset$^{2}$
\thanks{$^{1}$T. Wang and J. Whitman are with the Department of Mechanical Engineering, Carnegie Mellon University, Pittsburgh, PA 15213, USA.  {\{\tt\small tianyuw2, jwhitman\}@andrew.cmu.edu}}%
\thanks{$^{2}$M. Travers and H. Choset are with the Robotics Institute at Carnegie Mellon University, Pittsburgh, PA 15213, USA. {\{\tt\small mtravers, choset\}@cs.cmu.edu}}%
}

\begin{document}

\maketitle
\thispagestyle{empty}
\pagestyle{empty}

\begin{abstract}

Snake robots have the potential to maneuver through tightly packed and complex environments. 
One challenge in enabling them to do so is the complexity in determining how to coordinate their many degrees-of-freedom to create purposeful motion.
This is especially true in the types of terrains considered in this work: environments full of unmodeled features that even the best of maps would not capture, motivating us to develop closed-loop controls to react to those features.
To accomplish this, this work uses proprioceptive sensing, mainly the force information measured by the snake robot's joints, to react to unmodeled terrain.
We introduce a biologically-inspired strategy called directional compliance which modulates the effective stiffness of the robot so that it conforms to the terrain in some directions and resists in others.
We present a dynamical system that switches between modes of locomotion to handle situations in which the robot gets wedged or stuck.
This approach enables the snake robot to reliably traverse a planar peg array and an outdoor three-dimensional pile of rocks.

\end{abstract}

\vspace{-0.6em}
\section{Introduction}
\vspace{-0.3em}

\begin{figure}[t]
\vspace{0.65em}
\centering
\includegraphics[width=0.98\columnwidth]{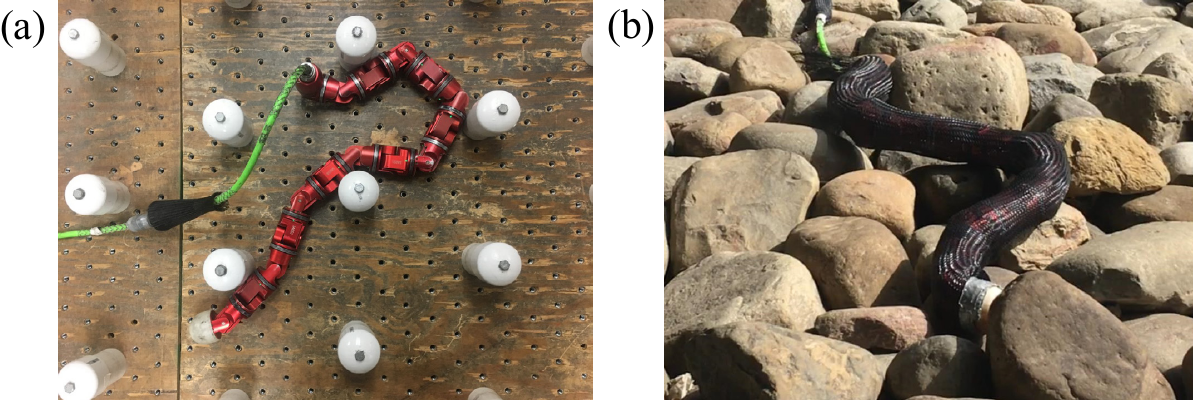}
\caption{In this work we present a control strategy to enable a snake robot to blindly locomote through unstructured terrains. 
(a) Indoors, in a planar peg board, the robot may be impeded when the posterior half of the robot becomes jammed between the pegs.
(b) Outdoors, while covered with a protective skin, the robot is impeded when its head becomes jammed between several rocks.
Our reactive control strategy infers that the robot is impeded and temporarily changes locomotive strategies.
}
\label{fig:stuck}
\vspace{-2em}
\end{figure}

Snake robots, composed of a chain of actuated joints, offer the potential to navigate challenging environments.
If the robot had perfect knowledge of its state and the environment, it might select which terrain features to exploit or avoid while locomoting through unstructured or confined environments \cite{kano2017tegotae,sanfilippo2017perception}. 
However, in practice, the robot has little prior knowledge of the terrain, making it prone to getting entangled in or jammed between obstacles (shown in Fig. \ref{fig:stuck}). 
This work presents a reactive controller which enables a snake robot to locomote through highly cluttered environments consistently using only proprioceptive (joint position and torque) sensors.

Prior work demonstrated that relatively straightforward reactive controllers based on proprioceptive feedback perform reasonably well in the types of terrain considered in this work \cite{travers2018shape,whitman2016shape}.
In these works, \textit{shape-based compliance} controllers couple the motion of the many joints to produce desired motions in the world, provide a reduced number of parameters to control, and adapt the body shape to conform to terrain features in the surrounding environment.
However, not all the terrain features are helpful for locomotion-- some serve as obstacles which impede the robot's movement rather than aiding it.
As a result, the robot's movements do not always result in desired propulsion. 
This work presents a novel means to infer when terrain features impede locomotion (and thus serve as obstacles), and adapts the shape-based compliance controller so that the robot moves through unstructured terrain more reliably. 

The approach to snake robot locomotion control presented in this paper centers around the concept of \textit{directional compliance}: allowing the robot to selectively admit forces applied on one side, and reject forces from the other side. 
This concept is inspired by the musculature and behavior of biological snakes \cite{jayne1988muscular,schiebel2019mechanical,astley2017side}.
We emulate this mechanism within the shape-based-compliance control strategy by selectively allowing some portions of the robot to comply to the environment, while others remain stiff to push off for forward propulsion.
We implement this mechanism within a dynamical system \cite{travers2016dynamical} that actively modulates the robot's local stiffness.

We empirically validate our approach both in a two-dimensional artificial indoor environment, a peg board, and in a three-dimensional outdoor environment, a rock pile. 
We experimentally compare our method to previous shape-based compliant control methods.
We find that our method enables the snake robot to locomote farther and more consistently in obstacle-rich environments.

The rest of this paper is organized as follows: Section \ref{sec:background} provides a background on related works. Section \ref{sec:method} describes our approach for snake robots navigating irregular terrain, which consists of inferring the presence of terrain features which obstruct forward progress using proprioceptive sensors, reacting to obstructions using directional compliance, and transitioning behaviors for robust locomotion. 
Section \ref{sec:experiment} experimentally validates our approach on a peg array and a rock pile and Section \ref{sec:results} compares its performance with the previous control strategy. Finally, conclusions and future works are discussed in Section \ref{sec:conclusion}.

\vspace{-0.5em}
\section{Background} \label{sec:background}
\vspace{-0.3em}

The reactive control strategy presented in this work is inspired by several existing methods that are reviewed in this section.


\vspace{-0.3em}
\subsection{Shape-based compliance}
\vspace{-0.3em}

The serpenoid curve, found throughout the snake robot literature \cite{hirose1993biologically}, is achieved by
\vspace{-0.5em}
\begin{equation}
    \theta_i = \kappa + A\sin(\eta s_i - \omega t),
    \label{eq:serpenoid}
\vspace{-0.5em}
\end{equation}
where $\theta_i$ is the $i$th joint angle, $\kappa$ the angular offset, $A$ the wave amplitude, $\omega$ the temporal frequency, and $t$ the time. $\eta$ is the spatial frequency, which determines the number of waves on the robot’s body. $s_i = i\delta s$ defines the position of the $i$th module along the robot backbone with respect to the head, where $\delta s$ is the distance between each joint. By tuning these parameters, various gaits, both biologically inspired and engineered, can be achieved \cite{tesch2009parameterized}.

To coordinate a snake robot's many internal degrees-of-freedom during locomotion, a shape function $h: \Sigma \rightarrow \mathbb{R}^N$ maps a point $\sigma$ in low-dimensional parameter space $\Sigma$ into the joint space $\theta \in \mathbb{R}^N$ of the $N$-link robot \cite{travers2018shape}. 
In this work, the amplitude $A$ of the serpenoid curve \eqref{eq:serpenoid} is updated dynamically during locomotion, while other parameters are fixed as constants, i.e., $\sigma = A$. 
The serpenoid curve serves as the shape function,
\vspace{-0.5em}
\begin{equation}
    h_i(\sigma) = h_i(A(t)) = \kappa + A(t)\sin(\eta s_i - \omega t).
    \label{eq:shapefunction}
\vspace{-0.5em}
\end{equation}

Shape-based compliant control \cite{travers2018shape} extends admittance control \cite{ott2010unified} to articulated locomotion, assigning spring-mass-damper-like dynamics to the shape parameters,
\vspace{-0.5em}
\begin{equation}
    M \ddot{\sigma}_d + B \dot{\sigma}_d + K (\sigma_d - \sigma_0) = F_\sigma,
    \label{eq:admittance}
\vspace{-0.5em}
\end{equation}
where $M$, $B$, and $K$ are positive-definite tuning matrices that govern the dynamic response of the desired shape parameters $\sigma_d$. The forcing term $F_\sigma$ is formed by transforming the external torques $\tau_{\text{ext}}$ measured by the joints into the shape-parameter space by
\vspace{-0.7em}
\begin{equation}
    F_\sigma = J\tau_{\text{ext}},
    \label{eq:shapeforce}
\vspace{-0.5em}
\end{equation}
given the Jacobian of the shape function $J = \frac{\partial h(\sigma)}{\partial \sigma}$. 
In the absence of external torques, the desired shape parameters $\sigma_d$ will converge back to the nominal shape parameters $\sigma_0$. 
The admittance control serves as a ``middle-level'' controller that takes feedback from the high-level planner (in the form of nominal shape values) and outputs commands for the low-level joint controllers (in the form of desired joint angle set-points).
In the remainder of this paper, we will refer to this form of shape-based compliance as ``nominal compliance'' to distinguish it from our proposed strategy, which augments this existing control strategy.

\vspace{-0.3em}
\subsection{Decentralized control with activation windows}\label{sec:window}
\vspace{-0.3em}

In irregular environments, recent work found that allowing different portions of the robot to independently conform in a decentralized manner resulted in more effective locomotion \cite{whitman2016shape}. 
This decentralization is accomplished by grouping sets of neighboring joints under ``activation windows.''
The joints contained within each activation window share shape parameters.
The activation windows are defined by travelling sigmoid functions, 
\vspace{-0.1em}
\begin{equation}
    \sigma(s,t) = \sum_{j=1}^W\sigma_j(t)\left[\frac{1}{1+e^{m(s_{j,l}-s)}} + \frac{1}{1+e^{m(s-s_{j,r})}}\right],
    \label{eq:window}
\vspace{-0.3em}
\end{equation}
where $W$ is the number of windows, $\sigma_{j}(t)$ is the shape parameter in window $j$, $m$ the sigmoid slope. 
Window $j$ spans the portion of robot backbone over $[s_{j,l},s_{j,r}] \subset [0,1]$. 
These activation windows move along the robot backbone with the travelling serpenoid curve,  which serves to pass spatial shape information down the body.

\vspace{-0.5em}
\section{Methods}\label{sec:method}
\vspace{-0.3em}

We divide our shape-based reactive control strategy into three parts: 
1) inferring the presence of terrain features that cannot be exploited by nominal compliance for propulsion (which for brevity, we will refer to as ``obstructions'') using only joint-level proprioceptions, 2) reacting actively to those obstructions and 3) transitioning between nominal and reactive behaviors. 

\vspace{-0.3em}
\subsection{Inferring the presence of obstructions}\label{sec:detection}
\vspace{-0.3em}

When the robot interacts with the terrain, contact forces cause external torques measured by the robot. 
We introduce a method to use these torque measurements to determine whether a portion of the body is stuck, jammed, or otherwise obstructed. 

When testing the shape-based compliance method of \cite{travers2018shape}, we observed that the robot frequently became wedged or stuck, and we sought trends within its internal state which were correlated with these instances. 
We found the amplitude shape parameters act as an indication of the robot's forward progression.
Each activation window, which groups a set of neighboring joints along the body, has its own amplitude shape parameter as per \eqref{eq:window}.
This means there are $W$ desired amplitudes for different portions of the robot $A_d = [A_{d,1}, ..., A_{d,W}]^T$.
Note that the amplitude $A_{d,j}$ of each activation window $j$ is independent of the number of joints contained within the window, and the number of windows is determined by the spatial frequency $\eta$ in \eqref{eq:shapefunction}. 
Each $A_{d,j}$ varies from its nominal value $A_{0,j}$. The difference between $A_{d,j}$ and $A_{0,j}$ reflects how much external torques have forced the robot to compliantly change its local body shape in the $j$th activation window.
To measure the degree to which each amplitude has varied, we take the mean value over a range of $k$ discrete time steps $t  = \{t_0,...,t_k\}$, 
\vspace{-0.5em}
\begin{equation}
    \mu_j = \frac{1}{k} \sum_{t = t_0}^{t_k}A_{d,j}[t].
    \label{eq:meanAmp}
\vspace{-0.5em}
\end{equation}
We observed that if a terrain feature aids the robot's locomotion, i.e., the robot moves through it without getting stuck, then $\mu_j \approx A_{0,j}$. 
That is, the mean of the amplitude remains near the nominal amplitude in the time period from when the robot makes contact with that feature and ends at the time when it loses contact.
On the other hand, if the $j$th window is obstructed by the terrain, the mean amplitude is offset from the nominal amplitude, i.e., $\Delta_j = |\mu_j - A_{0,j}| > 0$.

This observation on its own is insufficient to detect when the robots progress is impeded by its contact, however, since our snake robot has no external sensing and cannot directly sense when and where a contact begins or ends. 
Therefore we monitor the internal state by defining the \textit{shape absition} as the summation of the difference between the desired and the nominal shape parameter for the $j$th activation window over time,
\vspace{-0.7em}
\begin{equation}
    A_j^{\text{int}}[t_k] = \sum_{t=t_0}^{t_k}(A_{d,j}[t] - A_{0,j}) dt.
    \label{eq:absition}
\vspace{-0.7em}
\end{equation}
The $j$th amplitude absition $A_j^{\text{int}}$ remains near zero when locomotion progresses smoothly, but rapidly diverges if the robot's locomotion is hindered.
Thus, the shape absition serves as an indication of the robot's locomotion progress.
In practice, friction between the ground and the robot contributes a constant offset between $A_0$ and $A_d$ for each activation window. To remove the influence of the friction, we run several gait cycles on the ground to calibrate the offset and then subtract it from $A_j^{\text{int}}$. 

\begin{figure}[t]
\vspace{0.5em}
\centering
\includegraphics[width=0.9\columnwidth]{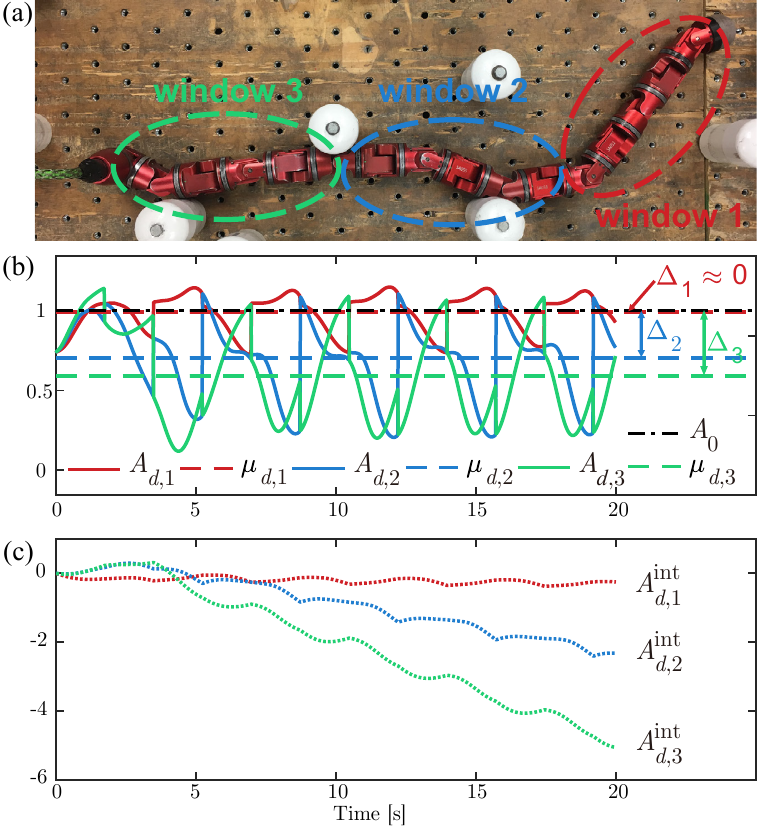}
\caption{We use joint position and torque feedback to infer when the robot's locomotion progress is obstructed. The amplitude of the body wave is allowed to vary over time in response to joint torques. 
(a) The snake robot has become jammed in place between three pegs around its mid-section and tail. 
(b) The desired amplitudes in the shape-based compliant controller vary over time for the three activation windows: the head (window 1, solid red), mid-section (window 2, solid blue) and tail (window 3, solid green). The means of amplitudes are shown for windows (dashed lines), and the difference of these means from the nominal amplitude of $A_0 = [1,1,1]^T$ (black dash-dotted line) are used to infer which windows are obstructed. 
(c) The resultant amplitude absitions for each window. The first window is not obstructed, so its absition remains near zero, but the other two windows are obstructed, so their absitions diverge from zero.}
\label{fig:absition}
\vspace{-1.5em}
\end{figure}

Fig. \ref{fig:absition} shows an example of the amplitudes and amplitude absitions when the robot's locomotion is impeded in mid-section and posterior windows. 
In this case, three activation windows are set along the robot's backbone.
In the anterior window, the offset $\Delta_1$ between the average desired amplitude $\mu_{d,1}$ and the corresponding nominal amplitude $A_{0,1}$ is approximately equal to zero, and as a result, the amplitude absition $A_1^\text{int}$ fluctuates about zero, suggesting the absence of obstructions near the head. 
However, $\Delta_2 > 0$, resulting in divergence of $A_2^\text{int}$ from zero, which indicates the presence of obstructions in the mid-section the robot.
Similarly, the divergence of $A_3^\text{int}$ from zero over this time period suggests the presence of obstructions around the posterior of the body. 

\vspace{-0.3em}
\subsection{Directional Compliance}\label{sec:DC}
\vspace{-0.3em}

Monitoring the shape absition serves as a means to infer obstructions to the locomotion progress while following a shape-based compliant controller.
To react to these obstructions we extend the nominal compliance to a new reactive strategy which we call \textit{directional compliance}. 
We are inspired by biological snakes that can actively tune the stiffness of selected muscles to propel against obstacles \cite{jayne1988muscular,schiebel2019mechanical,astley2017side}.
Directional compliance allows the snake robot to admit (comply to) forces exerted on one side of the body but reject forces exerted on the other side. 
We find this strategy enables the robot to actively react to, and move through, terrain features that impede their locomotion.

\subsubsection{Directional compliance in planar environments}

\begin{figure}[t]
\vspace{0.5em}
\centering
  \begin{subfigure}[b]{0.95\columnwidth}
    \includegraphics[width=\linewidth]{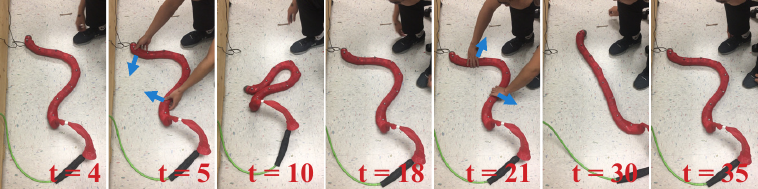}
    \label{fig:staticDC1}
  \end{subfigure}
\vspace{-0.8em}
    \begin{subfigure}[b]{0.9\columnwidth}
    \includegraphics[width=\linewidth]{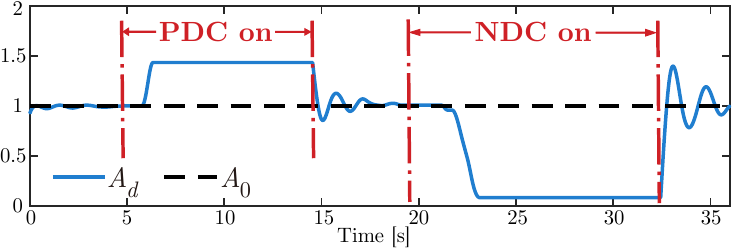}
    \label{fig:staticDC2}
  \end{subfigure}
\caption{Amplitude modulation of positive directional compliance (PDC) and negative directional compliance (NDC) were manually initiated in this demonstration, where control of the robot is fully centralized, i.e., there is one activation window that spans the whole body. 
At the start, a shape-based controller with nominal amplitude $A_0 = 1$ is used.
At $t = 5 s$, the controller is switched to PDC. 
The external forces exerted on the robot cause the amplitude $A_d$ to increase, and PDC prevents $A_d$ from decreasing.
At $t = 14 s$, the controller is switched back to nominal compliance, allowing $A_d$ to decrease and return to $A_0$. 
At $t = 19 s$, the controller is switched to NDC. External forces cause $A_d$ to decrease, then NDC prevents $A_d$ from increasing until at $t = 32$ the controller is switched back to NC. 
The robot then returns to $A_d = A_0$.}
\label{fig:staticDC}
\vspace{-1.5em}
\end{figure}

\begin{figure*}[t]
\vspace{0.1em}
\centering
 \includegraphics[width=0.91\textwidth]{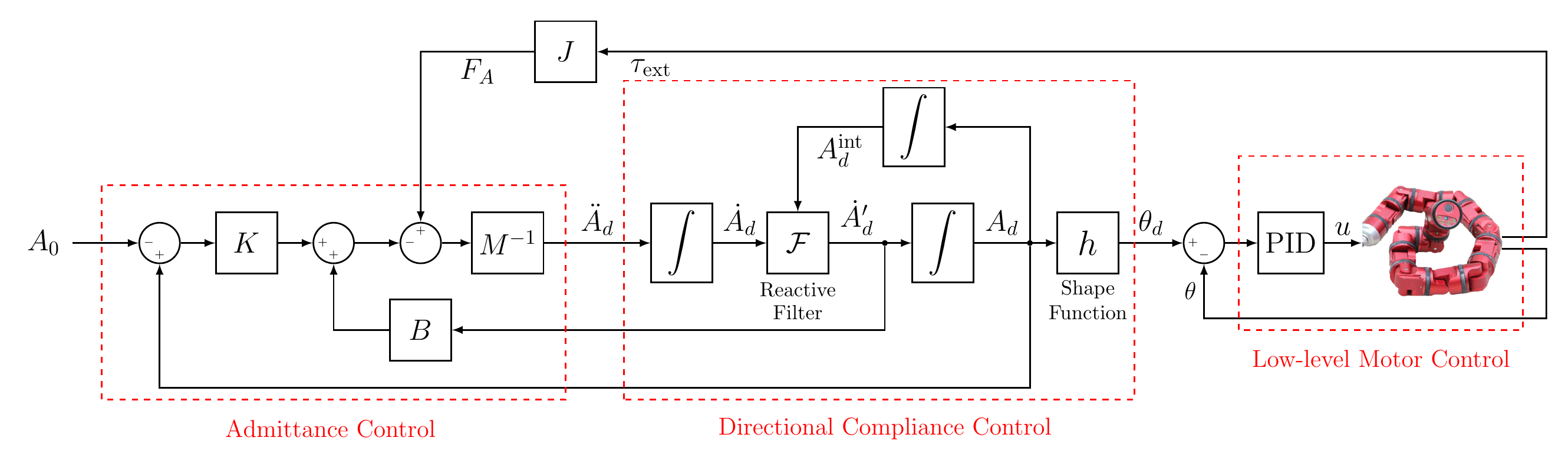}
\vspace{-1em}
\caption{Block diagram of the complete shape-based reactive control system with directional compliance.}
\label{fig:blockScheme}
\vspace{-1.5em}
\end{figure*}

The robot can increase its local curvature by increasing the amplitude in an activation window, or decrease local curvature by decreasing the amplitude. 
To make the robot directionally compliant, then, we can either admit forces resulting in increases in curvature, which we denote \textit{positive directional compliance}, or admit forces resulting in decreases in curvature, which we denote \textit{negative directional compliance}.
Recall that in nominal compliance (NC), we integrate the desired shape parameter $\ddot{\sigma}_d$ over time to dynamically vary the desired shape parameter $\sigma_d$.
We realize directional compliance by introducing a filter function $\mathcal{F}$ on $\dot{\sigma}_d$, such that the modified shape parameter derivative is $\dot\sigma_d' = \mathcal{F}(\dot\sigma_d)$. 
A filter that enables positive directional compliance (PDC) on the amplitude is
\vspace{-0.5em}
\begin{equation}
    \dot A_d'(t) = \mathcal{F}^+(\dot A_d(t)) = \left\{
    \begin{array}{ll}
    \dot A_d(t) & \text{if }\dot A_d(t) \geq 0,\\
    0 & \text{if }\dot A_d(t) < 0.
    \end{array} \right.
    \label{eq:PDC}
\vspace{-0.5em}
\end{equation}
Similarly, a filter that enables negative directional compliance (NDC) on the amplitude is
\vspace{-0.5em}
\begin{equation}
    \dot A_d'(t) = \mathcal{F}^-(\dot A_d(t)) = \left\{
    \begin{array}{ll}
    \dot A_d(t) & \text{if }\dot A_d(t) \leq 0,\\
    0 & \text{if }\dot A_d(t) > 0.
    \end{array} \right.
    \label{eq:NDC}
\vspace{-0.5em}
\end{equation}

The presence of an obstruction is detected by monitoring the amplitude absition $A^\text{int}$ as described in Section \ref{sec:detection}. 
When $A^\text{int}$ diverges from zero and $A^\text{int}<0$, we find that PDC, in which the amplitude is only allowed to increase locally, helps the robot propel itself forward through obstructions instead of passively conforming to them.
Similarly, NDC helps to overcome obstructions which cause $A^\text{int}$ to diverge from zero when  $A^\text{int}>0$.
See Fig. \ref{fig:staticDC} for an example of the effect of PDC and NDC on amplitude modulation.
A supplementary video of this demonstration can be found at \url{https://youtu.be/1BBPoczVgdY}.

\subsubsection{Directional compliance in three-dimensional environments}

Our snake robot has joints whose axes alternate between the dorsoventral and lateral directions.
For undulation in three-dimensional environments,
we generate two serpenoid curves on the dorsal and lateral planes by assigning them to odd and even joints, respectively,
\vspace{-0.3em}
\begin{equation}
    \begin{array}{l}
    \theta^o = A^o \sin(\eta^o s - \omega^o t) \\
    \theta^e = A^e \sin(\eta^e s - \omega^e t)
    \end{array} 
    \label{eq:3DAngle}
\vspace{-0.3em}
\end{equation}
where superscripts $o$ and $e$ represent odd and even joint indices. Note that planar locomotion can be achieved by setting $A^e = 0$. With \eqref{eq:3DAngle} as a three-dimensional shape function, we implement shape-based compliance separately on the dorsal and lateral plane. 
This allows the dorsoventral and lateral joints to conform independently to rough terrain.
The directional compliance filters \eqref{eq:PDC} and \eqref{eq:NDC} are applied independently to the amplitudes of the dorsoventral and lateral joints and windows.
With directional compliance in both planes, the robot is capable of reacting to three-dimensional irregularities that impede its locomotion.

\vspace{-0.3em}
\subsection{Dynamical system for behavior transition}\label{sec:transition}
\vspace{-0.3em}

\begin{figure*}[t]
\vspace{0.5em}
\centering
\includegraphics[width=0.9\textwidth]{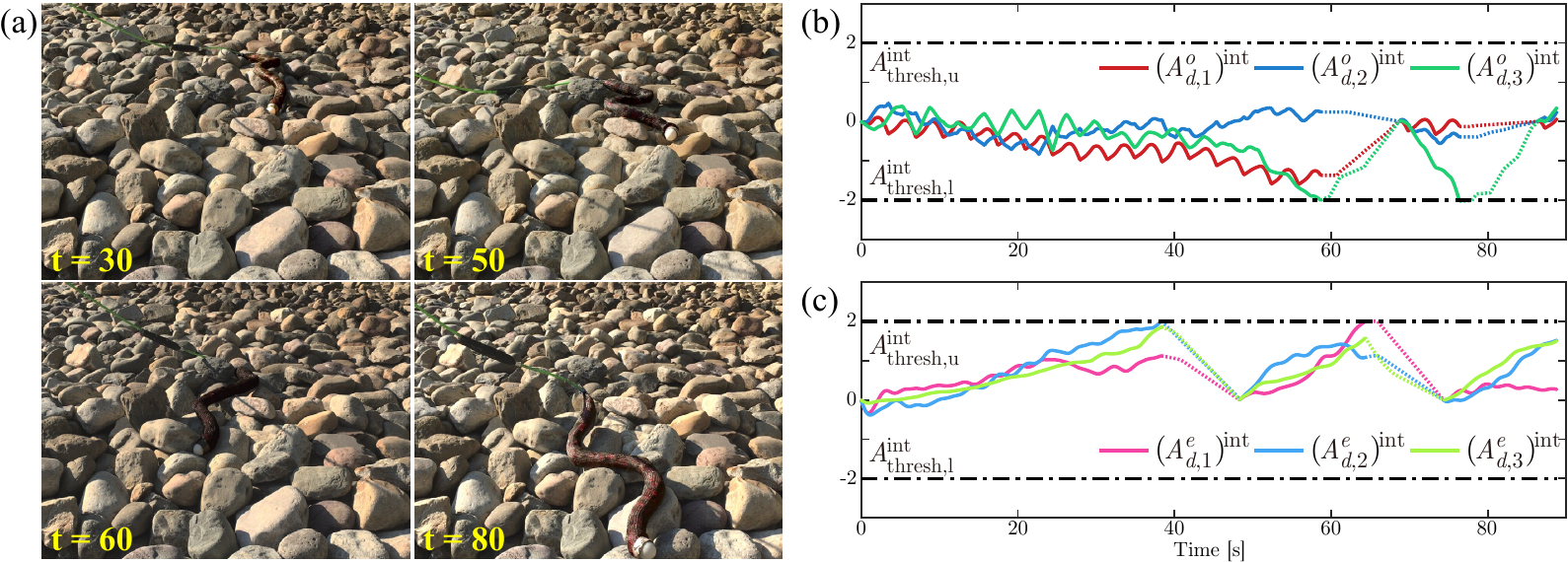}
\caption{Our control strategy enables a snake robot to blindly locomote over an unstructured outdoor rock pile.
The controller transitions between compliance modes: nominal compliance (NC), positive directional compliance (PDC) and negative directional compliance (NDC). (a) shows several selected frames of traversal over the rock pile. 
(b) shows the amplitude absitions in the dorsal plane for the anterior window (red), the mid-section window (blue) and the posterior window (green), and (c) shows amplitude absitions of three activation windows in the lateral plane (magenta, cyan and dark green, respectively). The black dash-dotted lines in both figures show upper and lower thresholds: when an amplitude absition (an integral of the amplitude deviation over time) $A^{\text{int}}$ reaches a threshold, then directional compliance is initiated until the amplitude absition returns to zero. These times are indicated here with dotted lines. 
In this example, the head of the robot is blocked by obstructions at $t = 38 s$ and $t = 65 s$. Then robot switches to NDC mode triggered by the lateral amplitude absition, and lifts its head portion to overcome the obstructions. 
PDC in the dorsal plane is also triggered twice at $t = 59 s$ and $t = 76 s$ to overcome obstructions located at sides of the body.
We find experimentally that this strategy results in significantly more consistent locomotive performance than does the nominal shape-based compliance.
}
\label{fig:rockpile}
\vspace{-1.0em}
\end{figure*}

Directional compliance provides a novel means for snake robots to overcome obstructions that impede their locomotion. 
Here we present a dynamical system that automatically transitions between the nominal compliance (NC), positive directional compliance (PDC), and negative directional compliance (NDC) behaviors based only on proprioceptive feedback.

The amplitude absition, as described in Section \ref{sec:detection}, is used to initiate the transition from nominal compliance to directional compliance.
This is accomplished by setting a threshold for the deviation of amplitude absition from zero, and initiating directional compliance based on the sign of the absition.
The upper and lower thresholds, $A_\text{thresh,u}^\text{int}$ and $A_\text{thresh,l}^\text{int}$, define the robot's sensitivity to obstructions, and are tuned by hand.
When directional compliance is enabled, these thresholds serve as new nominal amplitudes when calculating the amplitude absition.
When the amplitude absitions return to zero, the robot switches back to nominal compliance. 

A block diagram of the control system is shown in Fig. \ref{fig:blockScheme}. 
The nominal amplitude $A_0$, input by a user or high-level planner, defines the nominal shape that the robots takes in the absence of external forces.
The admittance control block uses joint torque feedback to compute a second derivative of the desired amplitude.
Our directional compliance block adds a filter on the first derivative of the desired amplitude to modulate the effective stiffness, and outputs joint angle set points.
A low-level PID controller embedded in each joint of the robot controls the actuators to follow the joint angle set points \cite{rollinson2014design}.
As the robot interacts with the environment, external forces from the terrain are sensed indirectly through joint torque measurements. 
The reactive filter $\mathcal{F}$ in Fig. \ref{fig:blockScheme} combines \eqref{eq:PDC} and \eqref{eq:NDC},
\vspace{-0.2em}
\begin{equation}
    \mathcal{F}(\dot A_d) = \left\{
    \begin{array}{llc}
    \dot A_d & \text{if }A_\text{thresh,l}^\text{int} \leq A_d^\text{int} \leq A_\text{thresh,u}^\text{int}, & \text{(NC)}\\
    \dot A_d & \text{if }A_d^\text{int} < A_\text{thresh,l}^\text{int} \text{ and } \dot A_d > 0, & \text{(PDC)}\\
    \dot A_d & \text{if }A_d^\text{int} > A_\text{thresh,u}^\text{int} \text{ and } \dot A_d < 0, & \text{(NDC)}\\
    0 & \text{otherwise}.
    \end{array} \right.
    \label{eq:fullFilter}
\vspace{-0.2em}
\end{equation}
NC is the default behavior when the shape absition $A_d^\text{int}$ lies between the upper and the lower thresholds. PDC is activated if $A_d^\text{int}$ drops below the lower threshold; the desired shape $A_d$ is only allowed to increase when negative $\dot{A}_d$ are filtered out. Similarly, NDC is activated if $A_d^\text{int}$ raises beyond the upper threshold; $A_d$ is only allowed to decrease when positive $\dot{A}_d$ are filtered out.
We found an effective threshold setting to be $A_\text{thresh,u}^\text{int} = 2A_0$ and $A_\text{thresh,l}^\text{int} = -2A_0$.

As described in Section \ref{sec:window}, the activation windows propagate from head to tail along the robot's backbone with the travelling wave of the serpenoid curve. 
When a new window is initiated at the head, it is set to the nominal amplitude $A_{1,d} = A_{0}$.
When a window passes off of the tail, its shape information is discarded, allowing the robot to leave behind the shape parameters with their associated terrain features.


\begin{figure*}[t]
\vspace{0.5em}
\centering
  \begin{subfigure}[b]{0.43\textwidth}
    \includegraphics[width=\linewidth]{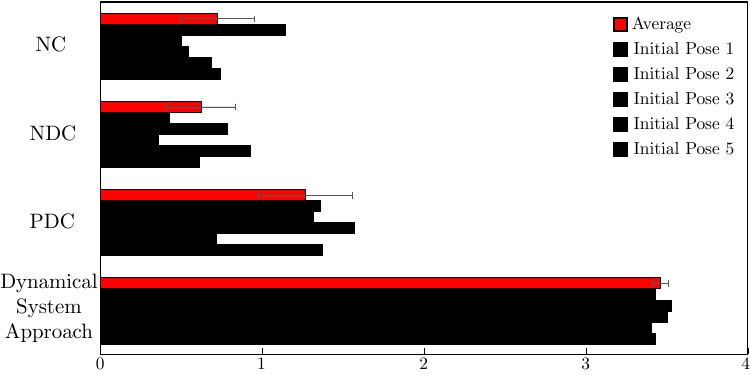}
    \caption{Velocity on peg array (body length/min)}
    \label{fig:expData1}
  \end{subfigure}
    \begin{subfigure}[b]{0.43\textwidth}
    \includegraphics[width=\linewidth]{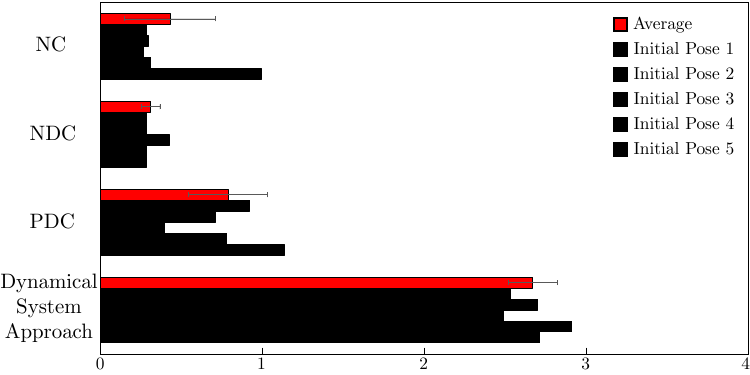}
    \caption{Velocity on rock pile (body length/min)}
    \label{fig:expData2}
  \end{subfigure}

\caption{A comparison of the performance of a nominal compliance controller (NC), negative directional compliance controller (NDC), positive directional compliance controller (PDC), and dynamical system for transitioning between these three controllers, (a) on a peg array and (b) over a rock pile. 
Each black bar represents the average velocity over the four trials for one controller in one initial pose. 
Each red bar depicts the average velocity of the controller for all trials in the environment, error bars indicating the standard deviation over the trials for each control strategy.}
\label{fig:expData} 
\vspace{-1.5em}
\end{figure*}

In summary, our reactive controller adds a biologically-inspired directional compliance layer to shape-based compliant control.
Directional compliance enables the robot to either push away from, or comply to, terrain features, based on the torque feedback measured over time. 
The amplitude of the body wave varies in response to external forces, and by tracking its change over time, we can infer the presence of obstructions and alter the compliance behavior accordingly.

Fig. \ref{fig:rockpile} depicts a sequence of video frames and the corresponding amplitude absitions of a successful traversal over an unstructured rock pile. 
Directional compliance in the dorsal and lateral plane are triggered independently by shape absitions in the two planes.

\vspace{-0.5em}
\section{Experiments}\label{sec:experiment}
\vspace{-0.3em}

We conducted an experimental comparison of the performance of shape-based compliant control variants.
All experiments were carried out with a snake robot composed of sixteen identical actuated joints \cite{rollinson2014design}. 
The joints were arranged such that the axes of rotation of neighboring modules were torsionally rotated ninety degrees relative to each other. Inside each actuator, a series-elastic element \cite{pratt1995series} provides joint torque feedback. 

Experiments were conducted in a randomly-distributed peg array and an unstructured outdoor rock pile.
For each environment, we compared four versions of shape-based control strategies:
\vspace{-0.3em}
\begin{itemize}
    \item Nominal compliance (NC)
    \item Negative directional compliance only (NDC)
    \item Positive directional compliance only (PDC)
    \item Dynamical system with NC, NDC, and PDC
\vspace{-0.3em}
\end{itemize}

The NC strategy is the shape-based controller from \cite{travers2018shape}, in which no filter is applied to the shape parameters derivatives. 
In the NDC strategy, a directional compliance filter in the form of $\mathcal{F}^-$ in \eqref{eq:NDC} is employed, which allows the amplitude to decrease in response to external forces, but not increase, in each window. 
In the PDC strategy, a directional compliance filter in the form of $\mathcal{F}^+$ in \eqref{eq:PDC} is employed, which allows the amplitude to increase but not decrease in each window.
The full dynamical system includes a filter in the form of \eqref{eq:fullFilter} which transitions between NC, NDC, and PDC, depending on trends in the amplitude over time in each window.

When operating within the peg array only odd modules are active, with three activation windows, where $A_0 = [\frac{\pi}{5}, \frac{\pi}{5}, \frac{\pi}{5}]^T$. 
When in the rock pile, three activation windows are assigned to both the dorsal and lateral plane, where $A_0^o = [\frac{\pi}{5}, \frac{\pi}{5}, \frac{\pi}{5}]^T$ and $A_0^e = [0, 0, 0]^T$.
Each controller was tested at five randomly selected initial positions and orientations.
For each initial pose, we collected four trials for each controller, starting from the same position, and averaged their displacements.
Controller performance in each condition was measured via the displacement of the geometric center of the robot after two minutes.

\vspace{-0.7em}
\section{Results}\label{sec:results}
\vspace{-0.3em}

We found that our full controller, a dynamical system transitioning between NC, NDC, and PDC, outperformed the previous method (NC) and the ablated variants (PDC and NDC).
The experimental results of the controllers' performances are displayed in Fig. \ref{fig:expData}.
We observed that NC is sensitive to the distribution of obstacles; it works well if the robot is placed in an initial position for which the nominal amplitude matches the distribution of terrain features, but frequently results in the robot becoming jammed between terrain features.
NDC performed worst, since it interferes with the robot's ability to push off of obstacles.
PDC behaves well initially, but does not produce robust locomotion.
This is due in part to the fact that when the amplitude can only increase, the robot has frequent self-collisions. 
The dynamical system strategy, which enables the robot to transition between the three control strategies, results in consistent planar and three-dimensional locomotion, and is the least sensitive to the robot's initial pose.
Example videos of the experiments can be found at \url{https://youtu.be/1BBPoczVgdY}.

\vspace{-0.2em}
\section{Conclusion}\label{sec:conclusion}
\vspace{-0.2em}

We presented a biologically-inspired dynamical system for snake robots to robustly navigate unmodeled obstacle-rich environments. 
We designed a decentralized method that, using only proprioceptive feedback, infers whether the robot is entangled in terrain features that impede rather than aid locomotion. 
Our directional compliance control strategy allows the robot to locally change the effective stiffness of each region of the body in order to overcome those obstacles.
We experimentally validated our approach in two different unstructured environments.

We found that directional compliance allows the robot to overcome common terrain features in outdoor environments, such as small bumps or small holes in the ground. 
However, this strategy is purely reactive, and cannot, for instance, transition to different modes of behavior needed to surpass larger terrain features like stairs.
In future work we will explore new motion patterns and control strategies for such scenarios.

This work focuses on one specific shape parameter in the serpenoid curve-- the amplitude. Future work will investigate the use of directional compliance in other shape parameters.
Further, we have so far assumed that the robot is fully blind, using no external sensing such as direct contact sensing, vision, or inertial sensors. A continued avenue for our research is how to include input from other sensing modalities into our reactive control strategies, as well as tighter integration between the mid-level controller and high-level planners. 

\section*{Acknowledgement}
We thank Dr. Henry Astley for providing the original idea behind this work.






\bibliographystyle{IEEEtran}
\bibliography{acc2020-DirectionalCompliance}

\end{document}